\pdfoutput=1
\documentclass[11pt]{article}

\let\Setlength\setlength % <--- Important: place this line before loading calc
\usepackage{calc}
\newlength{\arrayrulewidthOriginal}
\newcommand{\Cline}[2]{%
  \noalign{\global\Setlength{\arrayrulewidthOriginal}{\arrayrulewidth}}%
  \noalign{\global\Setlength{\arrayrulewidth}{#1}}\cline{#2}%
  \noalign{\global\Setlength{\arrayrulewidth}{\arrayrulewidthOriginal}}}

\usepackage{times}
\usepackage{latexsym}
\usepackage{graphicx}
\usepackage{amsmath}
\usepackage{multirow}
\usepackage{bm}

\usepackage[]{emnlp2021}

\usepackage{times}
\usepackage{latexsym}

\usepackage[T1]{fontenc}
\usepackage[utf8]{inputenc}

\usepackage{microtype}

\title{An Alignment-Agnostic Model for Chinese Text Error Correction}

\author{Liying Zheng \\
  \texttt{liying.zheng@outlook.com} \\\And
  Yue Deng \\
  \texttt{dengyue606@gmail.com} 
  \AND
  Weishun Song \\
  \texttt{songweishun474@gmail.com} \\\And
  Liang Xu \\
  \texttt{xlpaul@126.com} \\\And
  Jing Xiao \\
  \texttt{xiaojing661@pingan.com.cn} \\}

\begin{document}
\maketitle
\begin{abstract}
 This paper investigates how to correct Chinese text errors with types of mistaken, missing and redundant characters, which are common for Chinese native speakers. Most existing models based on detect-correct framework can correct mistaken characters, but cannot handle missing or redundant characters due to inconsistency between model inputs and outputs. Although Seq2Seq-based or sequence tagging methods provide solutions to the three error types and achieved relatively good results in English context, they do not perform well in Chinese context according to our experiments. In our work, we propose a novel alignment-agnostic detect-correct framework that can handle both text aligned and non-aligned situations and can serve as a cold start model when no annotation data are provided. Experimental results on three datasets demonstrate that our method is effective and achieves a better performance than most recent published models.
\end{abstract}

\section{Introduction}

Chinese text error correction plays an important role in many NLP related scenarios~\citep{martins2004spelling,afli2016using,wang2018hybrid,burstein1999automated}. For native Chinese speakers, common errors include mistaken characters, missing characters, and redundant characters. Mistaken characters refer to wrong characters needed to be replaced. Missing characters mean a lack of characters needed to be inserted into the identified position. Redundant characters mean useless or repeated characters needed to be deleted. Corrections for mistaken characters will not change the sentence length while corrections for the other two types will do. If texts only contain mistaken errors, we call it a text-aligned situation; if there exist missing or redundant errors, we call it a text non-aligned situation.

For text-aligned situation, many approaches apply the detect-correct framework, which is to detect the positions of wrong characters first and then correct them~\citep{hong2019faspell,zhang2020spelling,cheng2020spellgcn}. Despite of competitive performance of such methods, they cannot deal with text non-aligned situation with missing and redundant errors. For text non-aligned situations, the reversed order error or complex structural change with multiple errors are not in our scope, first because we target to cover common mistakes made by Chinese native speakers, which are different from foreign Chinese learners in Chinese error correction(GEC)~\citep{wang2020chinese,qiu2019twostage} task, second because the mentioned complex errors are beyond our model settings. The two mainstream model schemes for text non-aligned situation are Seq2Seq-based and sequence tagging-based. The former is inspired by machine translation, which sets wrong sentences as input and correct sentences as output~\citep{zhao2019improving,kaneko2020encoder,chollampatt2019cross,zhao2020maskgec,lichtarge2019corpora,ge2018reaching,junczys2018approaching}. Such approaches require a large number of training data and may generate uncontrollable results~\citep{kiyono2019empirical,koehn2017six}. The latter takes wrong sentences as input and modification operations of each token as output~\citep{awasthi2019parallel,malmi2019encode,omelianchuk2020gector}. However, as Chinese language has more than 20,000 characters that can generate many combinations of token operations, it is difficult for sequence tagging models to cover all combinations and generate results with high coverage rates.

To address the above issues, we propose an alignment-agnostic detect-correct model, which can not only handle text non-aligned errors compared to the current detect-correct methods, but also can relieve the problem of huge value search space leading to uncontrollable or low coveraged results of Seq2Seq or Sequence tagging based methods. We conduct experiments to compare our alignment-agnostic model with other models on three datasets: CGED 2020, SIGHAN 2015, SIGHAN-synthesize. Experimental results show that our model performs better than other models.

The contributions of our work include (1) proposal of a novel detect-correct architecture for Chinese text error correction, (2) empirical verification of the effectiveness of the alignment-agnostic model, (3) easy reproduction and fast adaptation to practical scenario with limited annotation data.

\section{Our Approach}
\subsection{Problem Description}
Chinese text error correction can be formalized as follows. Given a sequence of $n$ characters $X=(x_1,x_2,x_3,...,x_n)$ , the goal is to transform it into an $m$-character sequence $Y=(y_1,y_2,y_3,...,y_m)$, where $n$ and $m$ can be equal or not. The task can be viewed as a sequence transformation problem with a mapping function $f:X \rightarrow Y$

\begin{figure}[h]
\includegraphics[width=0.4\textwidth]{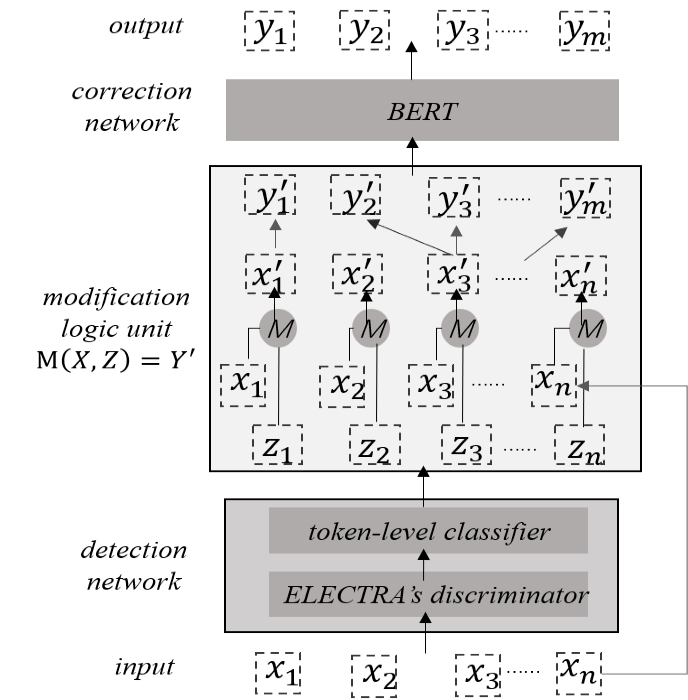}
\caption{Architecture of the alignment-agnostic model}
\label{figure1}
\end{figure}

%\vspace{-1em}
\subsection{Model}
As illustrated in Figure \ref{figure1}, the basic structure of our model includes a detection network evolved from ELECTRA discriminator~\citep{clark2020electra} and a correction network based on BERT MLM~\citep{devlin2018bert}. The two networks are connected through a modification logic unit and are trained separately. The detection network locates the errors and identifies error types. The modification logic unit handles where and how to correct. Finally the correction network focuses on detailed correction.

\textbf{The detection network} is composed of an ELECTRA discriminator and a token-level error type classifier. The architecture of ELECTRA discriminator has been described in \citet{clark2020electra}. Here we modify the original classifier, and define the new token-level classifier with four categories, namely $label_{keep}$, $label_{mistaken}$, $label_{missing}$, $label_{redundant}$. $label_{keep}$ means the character is correct and should not change. $label_{mistaken}$ indicates the character is mistaken and needs to be replaced. $label_{missing}$ denotes we should insert characters before the current character. $label_{redundant}$ means the character is useless and needs to be deleted. We get the label probability of each token with the 4-class token-level classifier:
\begin{equation}
    P_{i,label}(z_i=k|X)=softmax(w^Th^D(X))
\end{equation}
Where $P_{i,label}(z_i=k|X)$ denotes the conditional probability of character $x_i$ being tagged with the label $k$, $h^D(X)$ is the last hidden state of ELECTRA discriminator and $k$ is in label sets [$label_{keep}$, $label_{mistaken}$, $label_{missing}$, $label_{redundant}$]. The loss function of the detection network is:
\begin{equation}
    Loss_{detect}=-\sum_{i=1}^{n}\log{p_{i,label}}
\end{equation}

\textbf{The modification logic unit}, denoted by $M(X,Z)$, rewrites the input sequence $X$ according to detection network’s output $Z$:
%\vspace{-2.5mm}
\begin{equation}
if z_i=\left\{
\begin{aligned}
& label_{keep}, \; x_{i}^{'}=x_i \\
& label_{mistaken}, \; x_{i}^{'}=[MASK] \\
& label_{missing}, \; x_{i}^{'} = [MASK] \; x_i \\
& label_{redundant}, \; x_{i}^{'} =  ''
\end{aligned}
\right.
\end{equation}
%\vspace{-2.5mm}

Based on the above formula, we get a new sequence $X^{'}=(x_{1}^{'}, x_{2}^{'}, x_{3}^{'},..., x_{n}^{'},)$ . For each token with empty characters $''$, we delete it directly from the sequence $X^{'}$, For each token with $'[MASK] \; x_i'$, we reformulate it as two characters and obtain the final modified sequence $Y^{'}=(y_{1}^{'}, y_{2}^{'}, y_{3}^{'},..., y_{m}^{'},)$, whose length might be different from $X^{'}$. 

%\begin{center}
\begin{table*}[!h]
\centering
\begin{tabular}{|c|c|c|c|c|c|c|c|}
\hline
\multirow{2}{*}{\textbf{Test Set}} & \multirow{2}{*}{\textbf{Method}} & \multicolumn{3}{c|}{\textbf{Detection}} & \multicolumn{3}{c|}{\textbf{Correction}}\\ 
\cline{3-8}
~ & ~ & Prec. & Rec. & $F_1$. & Prec. & Rec. & $F_1$.\\
\hline
\multirow{7}{*}{SIGHAN 2015} & Hybrid~\citep{wang2018hybrid} & 56.6 & 69.4 & 62.3 & - & - & 57.1 \\
\cline{2-8}
~ & FASpell~\citep{hong2019faspell} & 67.6 & 60 & 63.5 & 66.6 & 59.1 & 62.6 \\
\cline{2-8}
~ & Confusionset~\citep{wang2018hybrid} & 66.8 & 73.1 & 69.8 & 71.5 & 59.5 & 64.9 \\
\cline{2-8}
~ & Soft-Masked BERT(2020) & 73.7 & \textbf{73.2} & \textbf{73.5} & 66.7 & \textbf{66.2} & 66.4 \\
\cline{2-8}
~ & our model(with a smaller training set) & \textbf{79.1} & 64.0 & 71.3 & \textbf{72.2} & 60.6 & \textbf{68.2} \\
\Cline{1.5pt}{2-8}
~ & SpellGCN~\citep{cheng2020spellgcn} & 74.8 & \textbf{80.7} & \textbf{77.7} & 72.1 & \textbf{77.7} & \textbf{75.9}\\
\cline{2-8}
~ & our model(with a larger training set) & \textbf{87.5} & 68.6 & 76.9 & \textbf{87.0} & 65.2 & 74.6\\
\hline
\end{tabular}
\caption{\label{table1} Performances of Different Methods on SIGHAN 2015. We trained our model on two datasets respectively.}
\end{table*}
%\end{center}
%\vspace{-2em}
\textbf{The correction network} is BERT. We do the prediction for positions with the $[MASK]$ symbol on the sequence $Y^{'}$.

\section{Experiments}
\subsection{Datasets and Metrics}
Chinese text error correction tasks mainly have two public datasets: the benchmark of SIGHAN 2015~\citep{tseng2015introduction} which only contains text-aligned data and the competition of CGED 2020~\citep{cged2020} which contains text non-aligned data. In order to better verify our models’ effectiveness on text non-aligned scenario, we synthesized some non-aligned data based on SIGHAN 2015 dataset. Next, we will introduce how to utilize the three datasets.

For SIGHAN 2015 dataset, in order to keep accordance with other models in comparison, we incorporated SIGHAN 2013 and 2014 datasets in the training phase, as well as the SIGHAN 2013 confusion set. The test set contains 1100 passages and the train set contains 8738 passages. To ensure comparability, we also trained another model on a considerably larger train set to be consistent with  SpellGCN'~\citep{cheng2020spellgcn}, which has 281379 passages in train set. We used the evaluation tool provided by SIGHAN, with metrics of precision (Prec.), recall(Rec.) and $F_1$, all are based on sentence level. 

CGED 2020 dataset is comprised of foreign Chinese learners' writing, and contains an additional error type besides the three types mentioned above, which is the reversed order. As this type happens less frequently in native Chinese writing scenario, and is also beyond the scope of our model setting, we remove 575 relevant samples from a total sample of 2586, and get 846 training samples and 1165 testing samples. In consequence, we redo experiments with  published models instead of comparing directly with the published benchmarks of other systems due to the inconsistency of test set.

To better verify our model’s effectiveness on text non-aligned scenario, we synthesized some non-aligned data based on SIGHAN 2015 dataset (SIGHAN-synthesized). For mistaken characters error type, we kept the original errors unchanged. For missing characters error type, we randomly selected 50\% samples and deleted one character from each of them. For redundant characters error type, we randomly selected 50\% samples and inserted characters in each of them through four ways. (1) We inserted repeated characters in 35\% of the selected samples. (2) We inserted confusing characters in 30\% of the selected samples. (3) We inserted characters from high-frequency words in 30\% of the selected samples. (4) We also inserted random characters in 5\% of the selected samples.

For CGED 2020 dataset and SIGHAN-synthesized dataset, we adopted the $M^2$ score~\citep{dahlmeier2012better} and ERRANT~\citep{bryant2017automatic} to evaluate models’ performance, which are two commonly used evaluation tools for text non-aligned situations.

%\begin{center}
\begin{table*}[htbp]
\centering
\begin{tabular}{|c|c|c|c|c|c|c|c|}
\hline
\multirow{2}{*}{\textbf{Test Set}} & \multirow{2}{*}{\textbf{Method}} & \multicolumn{3}{c|}{\textbf{\bm{$M_2$}(Correction)}} & \multicolumn{3}{c|}{\textbf{ERRANT(Correction)}}\\ 
\cline{3-8}
~ & ~ & Prec. & Rec. & $F_{0.5}$. & Prec. & Rec. & $F_{0.5}$.\\
\hline
\multirow{4}{*}{CGED 2020} & Copy-augmented(2019) & 4.62 & 0.8 & 2.36 & 3.51 & 0.56 & 1.7\\
\cline{2-8}
~ & Lasertagger(2019) & 14.99 & 3.48 & 9.02 & 12.95 & 2.61 & 7.22 \\
\cline{2-8}
~ & PIE(2019) & 22.3 & 10 & 17.9 & 17.1 & 6.6 & 13 \\
\cline{2-8}
~ & our model & \textbf{29.71} & \textbf{22.03} & \textbf{27.77} & \textbf{24.8} & \textbf{17.56} & \textbf{22.91} \\
\hline
\multirow{4}{*}{SIGHAN-synthesized} & Copy-augmented(2019) & 38.44 & 8.03 & 21.87 & 38.31 & 7.8 & 21.5 \\
\cline{2-8}
~ & Lasertagger(2019) & 51.29 & 43.21 & 49.44 & 50.14 & 39.99 & 47.72 \\
\cline{2-8}
~ & PIE(2019) & 54.1 & 47.6 & 52.6 & 52 & 42.6 & 49.8 \\
\cline{2-8}
~ & our model & \textbf{59.3} & \textbf{62.2} & \textbf{59.8} & \textbf{56.9} & \textbf{57.8} & \textbf{57} \\
\hline
\end{tabular}
\caption{\label{table2} The $M^2$ score and ERRANT score of Different Methods on CGED 2020 and SIGHAN-synthesized. }
\end{table*}
%\end{center}

%\begin{center}
\begin{table*}[htbp]
\centering
\begin{tabular}{|c|c|c|c|}
\hline
\multirow{2}{*}{\textbf{Method}} & \textbf{SIGHAN 2015} & \textbf{CGED 2020} & \textbf{SIGHAN-synthesized} \\ 
\cline{2-4}
~ & $F_1$ & $F_{0.5}$ & $F_{0.5}$ \\
\hline
ELECTRA+BERT & 38.7 & - & - \\
\hline
Finetune ELECTRA+BERT & 66.2 & 22.91 & 54.7 \\
\hline
Finetune ELECTRA+Finetune BERT & 68.2 & 22.54 & 54.4 \\
\hline
Finetune ELECTRA+Pretrain BERT & 42 & 22.6 & 57 \\
\hline
\end{tabular}
\caption{\label{table3} Ablation study of alignment-agnostic model on three datasets. }
\end{table*}
%\end{center}
%\vspace{-4.5em}

\subsection{Experiment Settings}
The pre-trained ELECTRA discriminator model and BERT model adopted in our experiments are all from \url{https://github.com/huggingface/transformers}. We use the large-size ELECTRA and the base-size BERT. We train detection network and correction network on the three datasets respectively by Adam optimizer with default hyperparameters. All experiments are conducted on 2 GPUs (Nvidia Tesla P100). 

For SIGHAN 2015, since it only contains one error type, we kept the default binary classifier of ELECTRA discriminator during finetuning. We applied two methods to retrain BERT. One is an unsupervised method by continue pretraining BERT with its original MLM objective. The other is a supervised method by masking mistaken characters and predicting them.

For CGED 2020 and SIGHAN-synthesized datasets, we added a 4-class classifier to recognize error types on ELECTRA discriminator’s last hidden layer and  finetune it. We applied the same methods as in SIGHAN 2015 to retrain BERT.

\subsection{Results}
Table~\ref{table1} shows the results on SIGHAN 2015 dataset. The first 5 lines implies that our method outperforms the method Soft-Masked BERT~\citep{zhang2020spelling} by 1.8\% on $F_1$ score in correction phrase. With a larger train set, our model achieved higher $F_1$ score in both detection and correction phases. Compared with the previous SOTA method  SpellGCN~\citep{cheng2020spellgcn}, our model showed higher precision and comparable $F_1$ score.

Table~\ref{table2} shows the results in comparison on  CGED 2020 dataset and SIGHAN-synthesized dataset. Our model performs the best on correction level, exceeding the second best model by 9.87\% on CGED 2020 and 7.2\% on SIGHAN-synthesized dataset with $F_{0.5} M^{2}$ score. Since Copy-augmented~\citep{zhao2019improving}, as a Seq2Seq model, requires a large size of training data to get an acceptable result, it underperforms Lasertagger~\citep{malmi2019encode} and PIE~\citep{awasthi2019parallel} models on both two datasets with a small training sample size. As analyzed before, sequence tagging models like Lasertagger and PIE do not work well on Chinese language due to huge value search space.
%\vspace{-0.3em}
\subsection{Ablation Study}
We carried out ablation study of our model on the three datasets. Table~\ref{table3} shows the results on correction level. For SIGHAN 2015,  finetuning ELECTRA can bing in a great improvement of 27.5\% on $F_1$ score, while finetuning BERT only generates a relatively small rise of 2\% on $F_1$ score and continue pretraining BERT leads to a decrease of 24.2\% on $F_1$ score. A possible reason is that finetuning can incorporate confusion sets knowledge about similar characters easy to be mistaken, while unsupervised pretraining may destroy the original learned words distribution when training data largely differs from the original ones. Besides, our model achieves 38.7\% on $F_1$ score with no training data and thus can work as a good baseline in cold start conditions. For CGED 2020 and SIGHAN-synthesized datasets, the two ways of retraining BERT didn't improve much. Compared with the results of other SOTA models, the modification and finetuning of ELECTRA is the most effective part.
%\vspace{-0.5em}
\section{Conclusion}
%\vspace{-0.5em}
We proposed a new detect-correct model for Chinese text error correction. It can handle both text-aligned and non-aligned situations, and can serve as a good baseline even in cold start situations. Experimental results on three datasets show that our model performs better than existing methods. Furthermore, it can be easily reproduced and achieve good results even with a small training data size, which is key to rapid application in the industry.

\bibliography{anthology,emnlp2021}
\bibliographystyle{acl_natbib}

\end{document}